\documentclass[letterpaper, 10 pt, conference]{ieeeconf}  

\IEEEoverridecommandlockouts                              

\usepackage{graphics} 
\usepackage{epsfig} 
\usepackage{mathptmx} 
\usepackage{times} 
\usepackage{amsmath} 
\usepackage{amssymb}  
\usepackage[caption=false]{subfig}
\usepackage{xcolor}
\title{\LARGE \bf
HIC-YOLOv5: Improved YOLOv5 For Small Object Detection
}

\author{\textbf{Shiyi Tang, Shu Zhang, Yini Fang} \\  
Heriot-Watt University, Ocean University of China, Hong Kong University of Science and Technology\\
\textit{st2015@hw.ac.uk, zhangshu@ouc.edu.cn, yfangba@connect.ust.hk}
\thanks{*This work was not supported by any organization}
\thanks{$^{1}$Shiyi Tang Author majors in Computer Science, Ocean University of China \& Heriot-Watt University 
       {\tt\small st2015@hw.ac.uk}}%
}

\begin{document}

\maketitle
\thispagestyle{empty}
\pagestyle{empty}

\begin{abstract}

Small object detection has been a challenging problem in the field of object detection. There has been some works that proposes improvements for this task, such as adding several attention blocks or changing the whole structure of feature fusion networks. However, the computation cost of these models is large, which makes deploying a real-time object detection system unfeasible, while leaving room for improvement. To this end, an improved YOLOv5 model: HIC-YOLOv5 is proposed to address the aforementioned problems.  Firstly, an additional prediction head specific to small objects is added to provide a higher-resolution feature map for better prediction. Secondly, an involution block is adopted between the backbone and neck to increase channel information of the feature map. Moreover, an attention mechanism named CBAM is applied at the end of the backbone, thus not only decreasing the computation cost compared with previous works but also emphasizing the important information in both channel and spatial domain. Our result shows that HIC-YOLOv5 has improved mAP@[.5:.95] by 6.42$\%$ and mAP@0.5 by 9.38$\%$ on VisDrone-2019-DET dataset.
\end{abstract}

\section{INTRODUCTION}
Object detection algorithm has been widely applied to smart systems of Unmanned Aerial Vehicles (UAVs), such as pedestrian detection and vehicle detection. It automates the analysis process of the photos taken by UAVs. However, the biggest issue of such applications lies in detecting small objects, as most of the objects in the photos become smaller from a higher altitude. This fact poses negative effects on the accuracy of object detection, including target occlusion, low target density, and dramatic changes in light. 
    
You Only Look Once (YOLO)\cite{redmon2016look}, a one-stage object detection algorithm, is dominating UAV systems due to its low latency and high accuracy. It takes an image as input and outputs the information of the objects in one stage. The lightweight model can achieve real-time object detection in UAV systems. However, it still has drawbacks in UAV scenarios with a large number of small objects. To address this issue, there have been previous works to improve the performance of small object detection. Some works \cite{result3}\cite{12112434} improves the whole structure of the feature fusion network of YOLOv5. Other works\cite{10.3389/fpls.2022.1091655} add several attention blocks in the backbone. However, the computation cost is large among the previous methods and there is still improvement space of the performance.

In this paper, we propose an improved YOLOv5 algorithm: HIC-YOLOv5 (Head, Involution and CBAM-YOLOv5) for small object detection, with better performance and less computation cost. We first add an additional prediction head---Small Object Detection Head (SODH) dedicated to detecting small objects from feature maps with a higher resolution. The features of tiny and small objects are more easily extracted when the resolution of the feature map increases. Secondly, we add a Channel feature fusion with involution (CFFI) between the backbone and the neck enhance the channel information, thereby improving overall performance. In this way, the performance is improved with more information transmitted to the deep network. Finally, we apply a lightweight Convolutional Block Attention Module (CBAM) at the end of the backbone, which not only has a lower computation cost than \cite{10.3389/fpls.2022.1091655} but also improves the total performance by emphasizing important channel and spatial features. The experiment result shows that our HIC-YOLOv5 has improved the performance of YOLOv5 on VisDrone dataset by 6.42$\%$(mAP@[.5:.95]) and 9.38$\%$(mAP@0.5).

Our main contributions can be summarized as follows:
    \begin{itemize}
    \item The additional prediction head is designed especially for small objects. It detects objects in higher-resolution feature maps, which contain more information about tiny and small objects.
    \item An involution block is added as a bridge between the backbone and neck to increase the channel information of the feature maps.
    \item CBAM is applied at the end of the backbone, thus more essential channel and spatial information is extracted while the redundant ones are ignored.
    \end{itemize}

\section{RELATED WORKS}

\subsection{Object Detection Primer}

The main purpose of object detection is to locate and classify objects in images, in the form of bounding boxes and confidence scores labeled on the objects. There are two types of object detection pipelines: two-stage and one-stage detectors. The two-stage detectors (e.g., R-CNN\cite{DBLP:journals/corr/GirshickDDM13}, SPP-net\cite{DBLP:journals/corr/HeZR014}, Fast R-CNN\cite{DBLP:journals/corr/Girshick15} and FPN\cite{DBLP:journals/corr/LinDGHHB16}) first generate region proposals, and then apply object classification and height and width regression. The one-stage detectors (e.g., YOLO series and SSD\cite{Liu_2016}) use a deep learning model, which directly takes an image as input and outputs bounding box coordinates and class probabilities. 

\begin{figure}
\centering
\includegraphics[width=\linewidth]{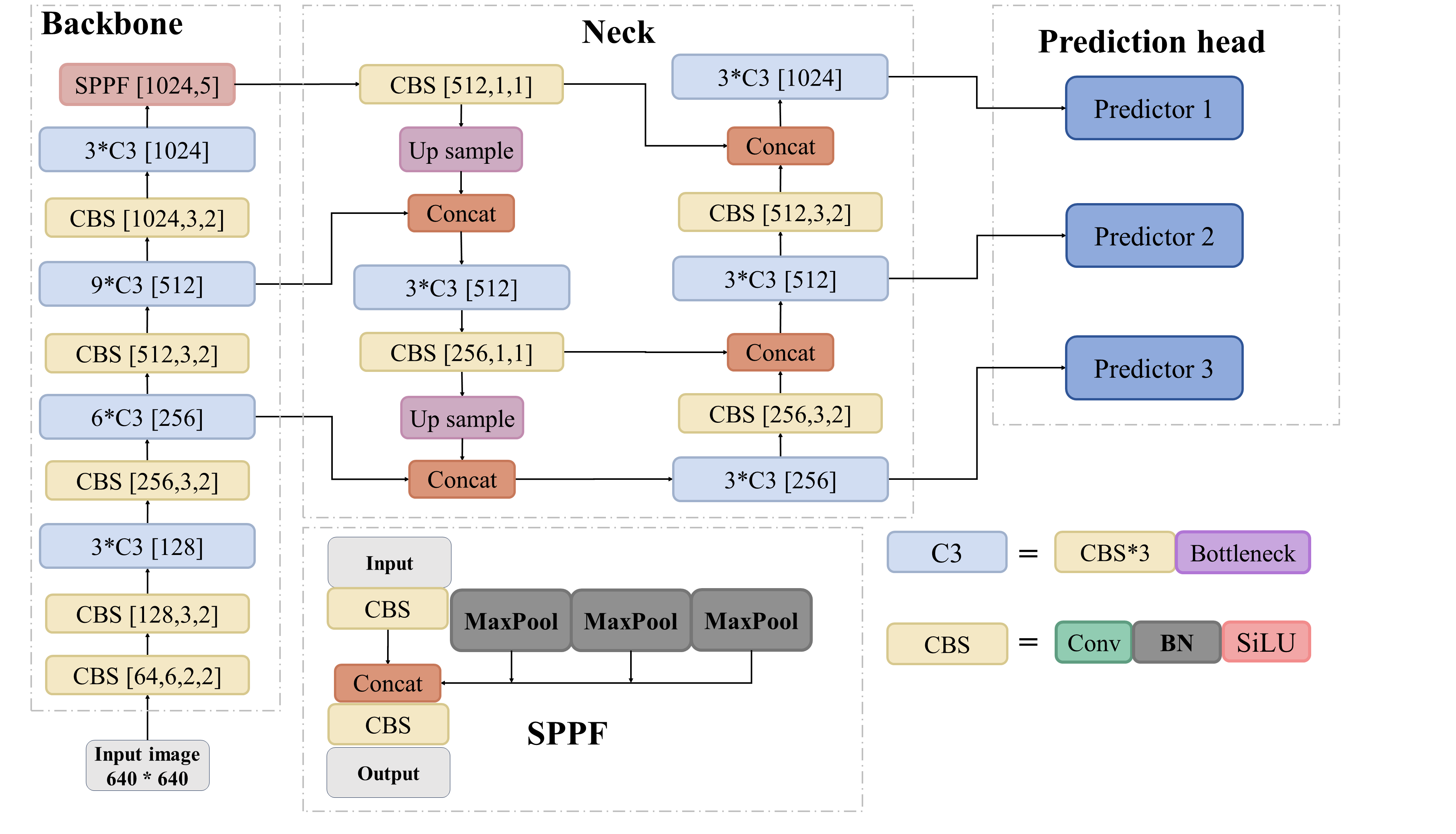}
\caption{Structure of YOLOv5-6.0.}
\label{fig:default yolov5}
\end{figure}

Among all the YOLO series, YOLOv5 is the most suitable algorithm for real-time object detection due to its promising performance and excellent computational efficiency. There have been several versions of YOLOv5, with the same main structure but a few differences on some small modules. In this paper, we choose to use YOLOv5-6.0 as our experimental algorithm. The main structure of YOLOv5-6.0 are shown in Fig. \ref{fig:default yolov5}.


The backbone of YOLOv5 firstly extracts features from the input image and generates different sizes of feature maps. These feature maps are then fused with the feature maps in the neck. Finally, three different feature maps generated from the neck are sent to the prediction head. The detailed information is described as follows:

The backbone includes several Conv, CSPDarkNet53 (C3), and SPPF modules. The Conv module adopts Conv2d, Batch Normalization, and SiLU activation function. C3, which is based on the CSPNet\cite{wang2019cspnet}, is the main module to learn the residual features. It includes two branches: one branch adopts three Conv modules and several Bottlenecks, and another branch only uses one Conv module. The two branches are finally concatenated together and fed into the next modules. SPPF modules are added at the end of the backbone, which is an improved type of Spatial Pyramid Pooling (SPP)\cite{DBLP:journals/corr/HeZR014}. It replaces the large-sized pooling kernels with several cascaded small-sized pooling kernels, aiming to increase the computation speed while maintaining the original function of integrating feature maps of different receptive fields to enrich the expression ability of features.

\begin{figure}
\centering
\includegraphics[width=\linewidth]{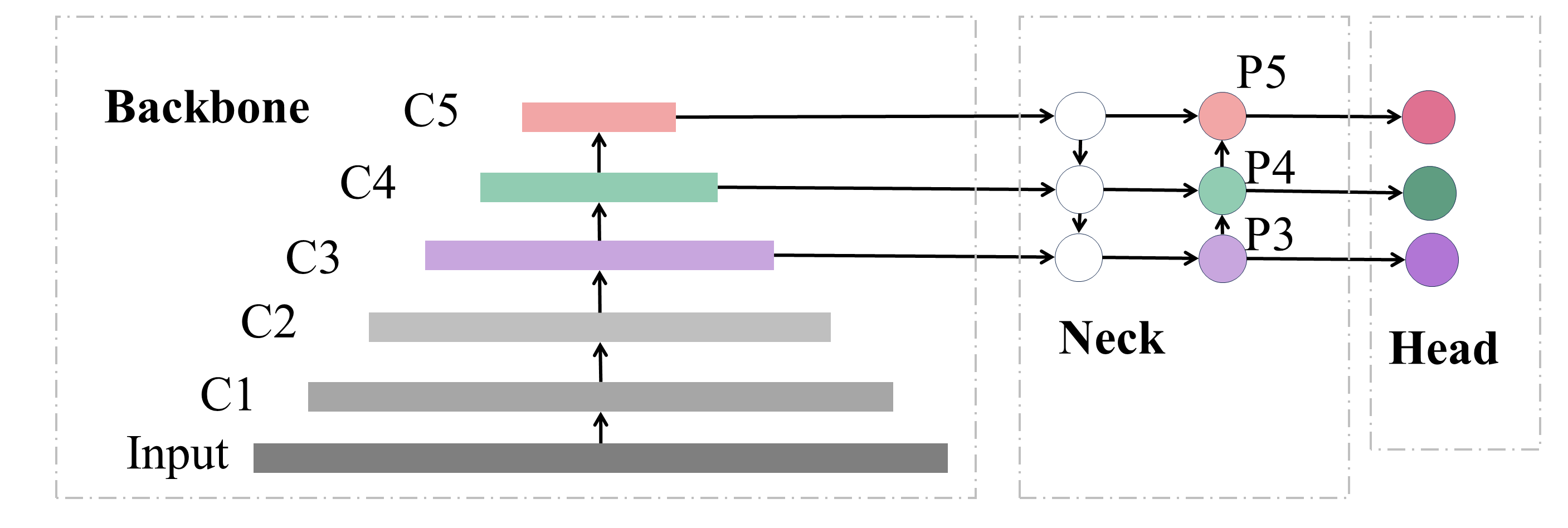}
\caption{FPN and PANet structure in YOLOv5.}
\label{fig:FPN&PANet}
\end{figure}

The neck of YOLOv5 draws on the structure of the Feature Pyramid Network (FPN) and Path Aggregation Network (PANet). The structure of FPN and PANet in YOLOv5 are shown in Fig. \ref{fig:FPN&PANet}. FPN mainly consists of two paths: Bottom-up and Top-down. Bottom-Up path responds to the backbone of YOLOv5, which gradually decreases the size of the feature map to increase the semantic information. Top-down path takes the smallest feature map generated by Bottom-up path as input and gradually increases the size of the feature map using an upsample, thus increasing the semantic information of low-level features. Finally, feature maps with the same size in the two paths are laterally connected together to increase the semantic representation on multiple scales. PANet adds a Bottom-up path based on FPN. Therefore, the position information at the low level can also be transmitted to the deep level, thus enhancing the positioning ability at multiple scales.

There are totally 3 prediction heads in YOLOv5, which aims to detect three sizes (80 $\times$ 80, 40 $\times$ 40, 20 $\times$ 20) of objects respectively: large, medium and small, where the image resolution is 640 $\times$ 640. The prediction heads divide grids on these three feature maps according to the dimensions of the feature maps. Then, three groups of anchors with different aspect ratios for each grid on each feature map are set to generate candidate bounding boxes. Finally, Non-Maximum Suppression (NMS) is applied to discard the overlapping bounding boxes and output the final bounding boxes, which include the locations and sizes of the boxes, and the confidence scores of the objects.

\subsection{Previous Works on Small Object Detection}
There have been numerous prior efforts aimed at enhancing the detection performance of small objects. Certain studies have focused on optimizing the overall architecture of the neck component within YOLOv5. For instance, \cite{12112434} replaces the PANet in YOLOv5 with a weighted bidirectional feature pyramid Mul-BiFPN and \cite{result3} introduces a new feature fusion method PB-FPN to the neck of YOLOv5. However, both methods choose to change the entire structure of neck to achieve better feature fusion, which result in larger computation cost. Instead, we introduce a lightweight involution block between the backbone and the neck, aiming to improve the performance of PANet in the neck of YOLOv5 with less computation cost and higher accuracy. \cite{result2} introduces a spatio-temporal interaction module, which applies recursive gated convolution to make greater spatial interaction, but causes channel information loss because of 1 $\times$ 1 convolution layers. In our structure, the involution block can effectively address this problem. Moreover, some works try to apply attention mechanisms\cite{woo2018cbam}\cite{dosovitskiy2021image}. Attention mechanisms have been widely applied in the field of computer vision, which learns to emphasize essential parts and ignore the unimportant ones in an image. There are various types of attention mechanisms, such as channel attention, spatial attention, temporal attention, and branch attention. There have been previous works that integrate transformer layers into YOLOv5. \cite{10.3389/fpls.2022.1091655} adds a transformer layer at the end of the backbone. However, it requires large computation costs and is difficult to train the model when the input image size is large. Compared with these methods, we add a lightweight CBAM block at the end of the backbone, which aims to use less computation cost and focus more on essential information when extracting features.

\section{METHODOLOGY}

\begin{figure}
\centering
\includegraphics[width=\linewidth]{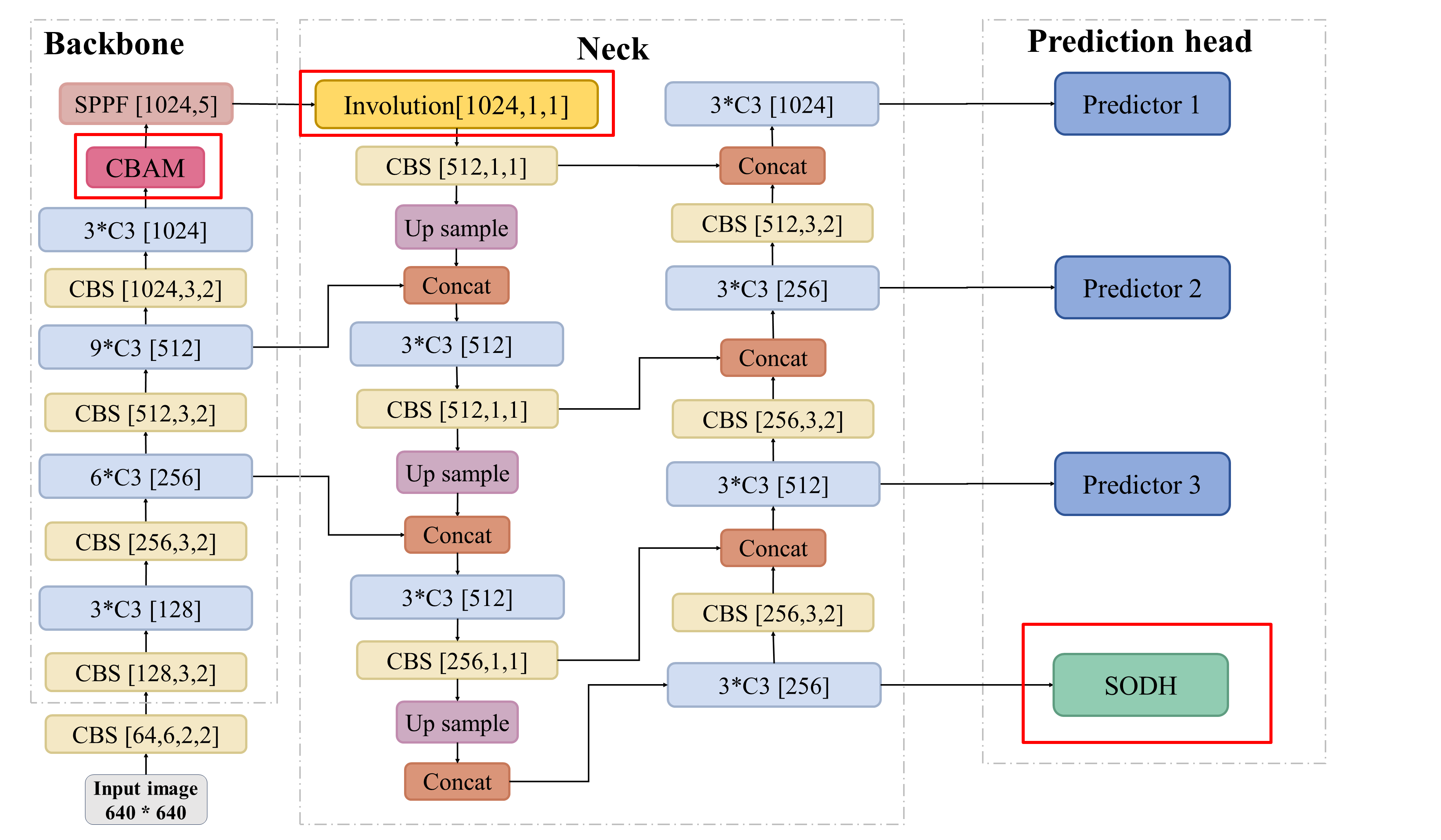}
\caption{Structure of HIC-YOLOv5.}
\label{fig:HIC-yolov5}
\end{figure}

The structure of HIC-YOLOv5 is shown in Fig. \ref{fig:HIC-yolov5}. Original YOLOv5 consists of 3 sections: backbone for feature extraction, neck for feature fusion, and 3 prediction heads. Based on the default model, we propose three modifications: 1) we add an additional prediction head to detect layers with high-resolution feature maps for small and tiny objects specifically; 2) an Involution block is adopted at the beginning of the neck to improve the performance of PANet; 3) we incorporate the Convolutional Block Attention Module (CBAM) into the backbone network.

\subsection{Convolutional Block Attention Module(CBAM)}

Previous works add CBAM into the Neck block when generating feature pyramids. However, the parameters and computing cost increase because some feature maps connected with CBAM have large sizes. Moreover, the model is difficult to train due to the large amount of parameters. Hence, we adopt CBAM in the Backbone network with the purpose of highlighting significant features when extracting features in the backbone, rather than generating feature pyramids in the neck. Moreover, the feature map size as the input of CBAM is only 20 $\times$ 20 which is 32 times smaller than the 640 $\times$ 640 full image  so that the computing cost will not be large.

\begin{figure}[ht]
\centering
\includegraphics[width=0.7\linewidth]{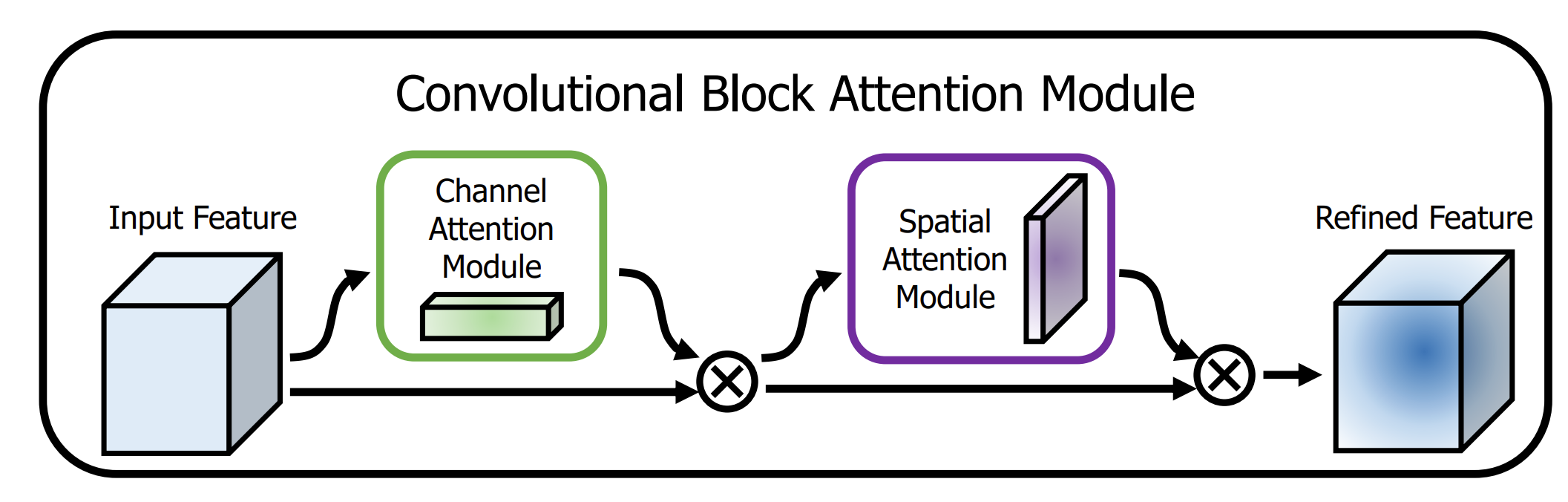}
\caption{Structure of CBAM\cite{woo2018cbam}.}
\label{fig:cbam}
\end{figure}

CBAM is an effective model based on attention mechanism, which can be conveniently integrated into CNN architectures. It consists of 2 blocks: Channel Attention Module and Spatial Attention Module, as shown in Fig. \ref{fig:cbam}. The two modules respectively generate a channel and a spatial attention map, which are then multiplied with the input feature map to facilitate adaptive feature refinement. Therefore, the meaningful features along both channel and spatial axes are emphasized, while the redundant ones are suppressed. The Channel Attention Module performs global Max-pooling and Average-pooling for feature maps on different channels and then executes element-wise summation and sigmoid activation. The Spatial Attention Module performs a global Max-pooling and Average-pooling for Values of pixels in the same position on different feature maps and then concatenates the two feature maps, followed by a Conv2d operation and sigmoid activation.

 \subsection{Channel Feature Fusion with Involution (CFFI)}
The neck of YOLOv5 adopts PANet, which introduces the bottom-up path augmentation structure on the basis of FPN. The corresponding structure of FPN and bottom-up path augmentation in YOLOv5 is shown in Fig. \ref{fig:FPN&PANet}. Particularly, FPN has great ability to detect small and tiny targets by fusing features of high and low layers so as to obtain high resolution and strong semantics features. However, a 1 $\times$ 1 convolution is adopted to reduce the number of channels at the beginning of the neck in original YOLOv5, where the calculation efficiency is significantly improved, but the channel information is also reduced, leading to poor performance of PANet. Inspired by \cite{10.3389/fpls.2022.1091655},  we add an Involution block between the backbone and the neck. The channel information is improved and shared, resulting in the reduction of information loss during the initial phases of FPN. As a result, this improvement contributes to the enhanced performance of FPN, particularly benefiting the detection of objects with smaller sizes. Moreover, it is emphasized that Involution has better adaptation to various visual patterns in terms of different spatial positions.\cite{li2021involution}

\begin{figure}
\centering
\includegraphics[width=0.6\linewidth]{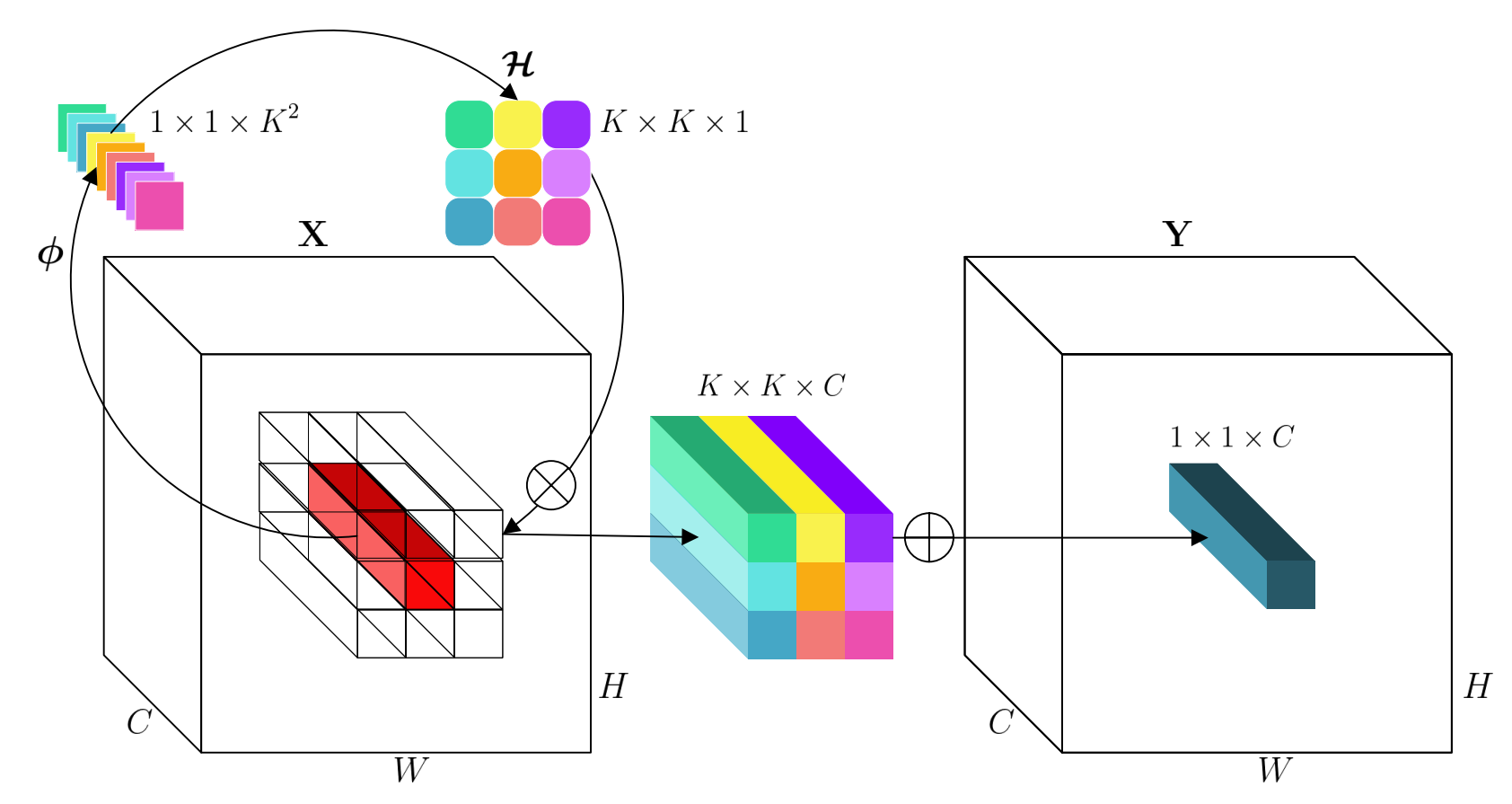}
\caption{Structure of Involution block\cite{li2021involution}.}
\label{fig:involution}
\end{figure}

The structure of Involution is illustrated in Fig. \ref{fig:involution}. Involution kernels, represented as $\mathcal{H} \in \mathbb{R}^{H \times W \times K \times K \times G}$, are designed to incorporate transformations that exhibit inverse attributes in both the spatial and channel domains, where $H$ and $W$ represents the height and width of the feature map, $K$ is the kernel size and $G$ represents the number of groups, where each group shares the same involution kernel. Particularly, a specific involution kernel, denoted as $\mathcal{H}_{i, j, \cdot, \cdot, g} \in \mathbb{R}^{K \times K}, g=1,2,...,G$, is designed for the pixel $\mathbf{X}_{i, j} \in \mathbb{R}^C$ (the subscript of C is omitted for brevity), while being shared across the channels. Finally, the output feature map of involution $\mathbf{Y}_{i, j, k}$, is obtained as follows: 
\begin{equation}
\mathbf{Y}_{i, j, k}=\sum_{(u, v) \in \Delta_K} \mathcal{H}_{i, j, u+\lfloor K / 2\rfloor, v+\lfloor K / 2\rfloor,\lceil k G / C\rceil} \mathbf{X}_{i+u, j+v, k}
\end{equation}
Therefore, the information contained in the channel dimension of a single pixel is implicitly dispersed to its spatial vicinity, which is useful to obtain the enriched receptive field information.

\subsection{Prediction Head}
The different resolutions (80 $\times$ 80, 40 $\times$ 40, and 20 $\times$ 20) of 3 prediction heads in YOLOv5 make a great contribution to the detection ability in various application scenarios, but also make it difficult to detect small and tiny objects. The reason why the performance of YOLOv5 on tiny object detection is poor is that the features of tiny objects which only contain few pixels are likely to be ignored. Although convolutional blocks play an important role in extracting the features from feature maps, they also reduce the resolution of feature maps when the depth of the network increases, thus the features of the tiny object are difficult to extract. In order to solve this issue and inspired by  \cite{12112434}\cite{zhu2021tphyolov5}\cite{result2}, we propose an additional prediction head---Small Object Detection Head (SODH), which aims to detect feature maps with larger resolution (160 $\times$ 160). It becomes increasingly effortless to extract features from small and minuscule objects.

Each prediction head takes the feature extracted and fused by backbone and neck as input, and finally outputs a vector, which consists of the regression bounding box (coordinate and size), the confidence of the object's border and the class of the object. Before generating the final bounding boxes, we generate anchors to form the candidate bounding boxes. These anchors are generated by k-means according to the dataset and are defined in 3 different scales for the 3 prediction heads, adapting to small, middle and large objects respectively. Anchors of the additional prediction head are also generated by k-means.

\subsection{Loss Function}
The loss function of HIC-YOLOv5  consists of three sections: objectness, bounding box and class probability, which can be represented as follows:
\begin{equation}
\operatorname{Loss}=\alpha \operatorname{Loss}_{o b j}+\beta\text { Loss }_{b o x}+\gamma \operatorname{Loss}_{c l s}
\end{equation}

We use binary cross entropy loss for both objectness and class probability, and CIoU loss \cite{zheng2021enhancing} for bounding box regression.

\subsection{Data Augmentation}
Data augmentation is an essential technique to enhance the robustness of the model. In YOLOv5, it includes Mosaic, Copy paste, Random affine, MixUp, HSV augmentation and Cutout. Except that, we found that many small people and cars are in the center of a picture Visdrone2019. Therefore we add extra center cropping to the data augmentation techniques mentioned above..

\section{EXPERIMENTAL RESULTS}

The experiments were conducted on the VisDrone2019 dataset, and the obtained experimental results demonstrate that our proposed YOLOv5 exhibits excellent performance in terms of detection accuracy.

\subsection{Experimental Setting}
\subsubsection{Experimental Equipment}
In this experiment, the CPU is 15 vCPU Intel® X Platinum 8358P CPU @ 2.60GHz, the GPU is NVIDIA A40 with 48 GB of Graphics memory. The algorithm is implemented by PyTorch, using CUDA 11.6 to operation acceleration.
\subsubsection{Dataset}
The dataset used in this experiment is VisDrone2019, which is a comprehensive benchmark facilitating the integration of drone technology and visual perception. VisDrone2019 was collected by the AISKYEYE team at Lab of Machine Learning and Data Mining, Tianjin University, China. It comprises 288 video clips consisting of 261,908 frames and 10,209 static images captured by diverse drone-mounted cameras across different locations separated by thousands of kilometers in China, environments containing both urban and rural, objects including pedestrians, vehicles, bicycles etc., and densities from sparse to crowded scenes. Notably, this dataset was acquired using multiple drone platforms with varying models under different scenarios as well as weather and lighting conditions. The dataset is divided into a training set, a validation set and a testing set, with 6471, 548, 1610 images respectively.

\begin{figure}
    \centering
    \subfloat[Instances. Information of 10 classes in VisDrone-2019.]{
        \label{instances}\includegraphics[width=0.2\textwidth]{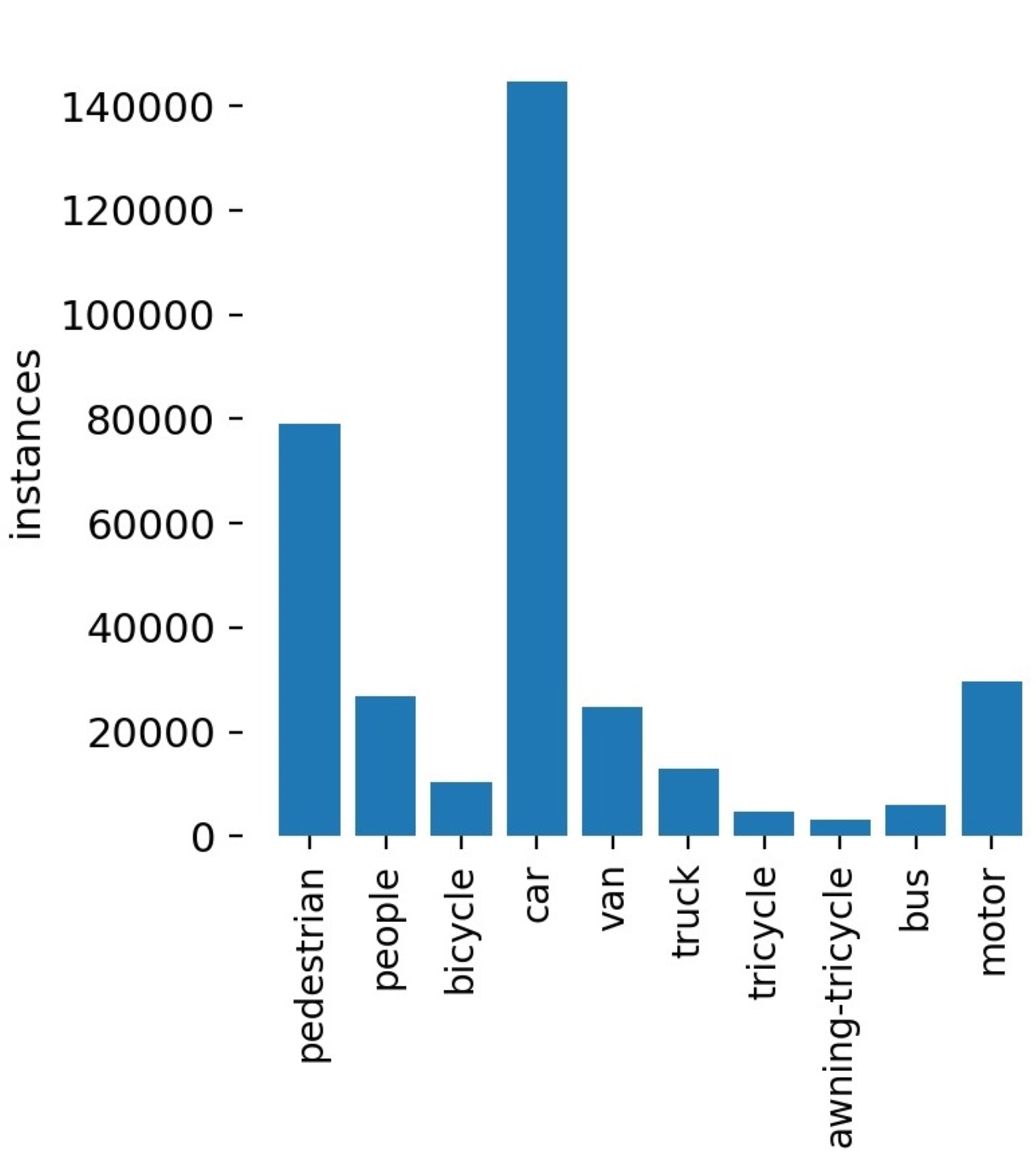}
    }
    \quad
    \subfloat[Locations. The locations of objects in an image, whose height and width are assumed to be 1.]{
        \label{location}\includegraphics[width=0.2\textwidth]{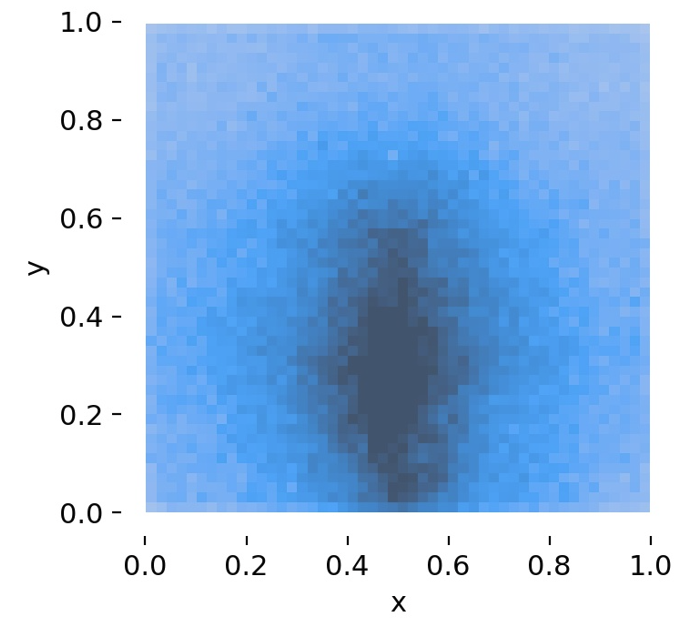}
    }

    \subfloat[Distribution. The distribution of object areas in VisDrone-2019.]{
        \label{sizes}\includegraphics[width=0.28\textwidth]{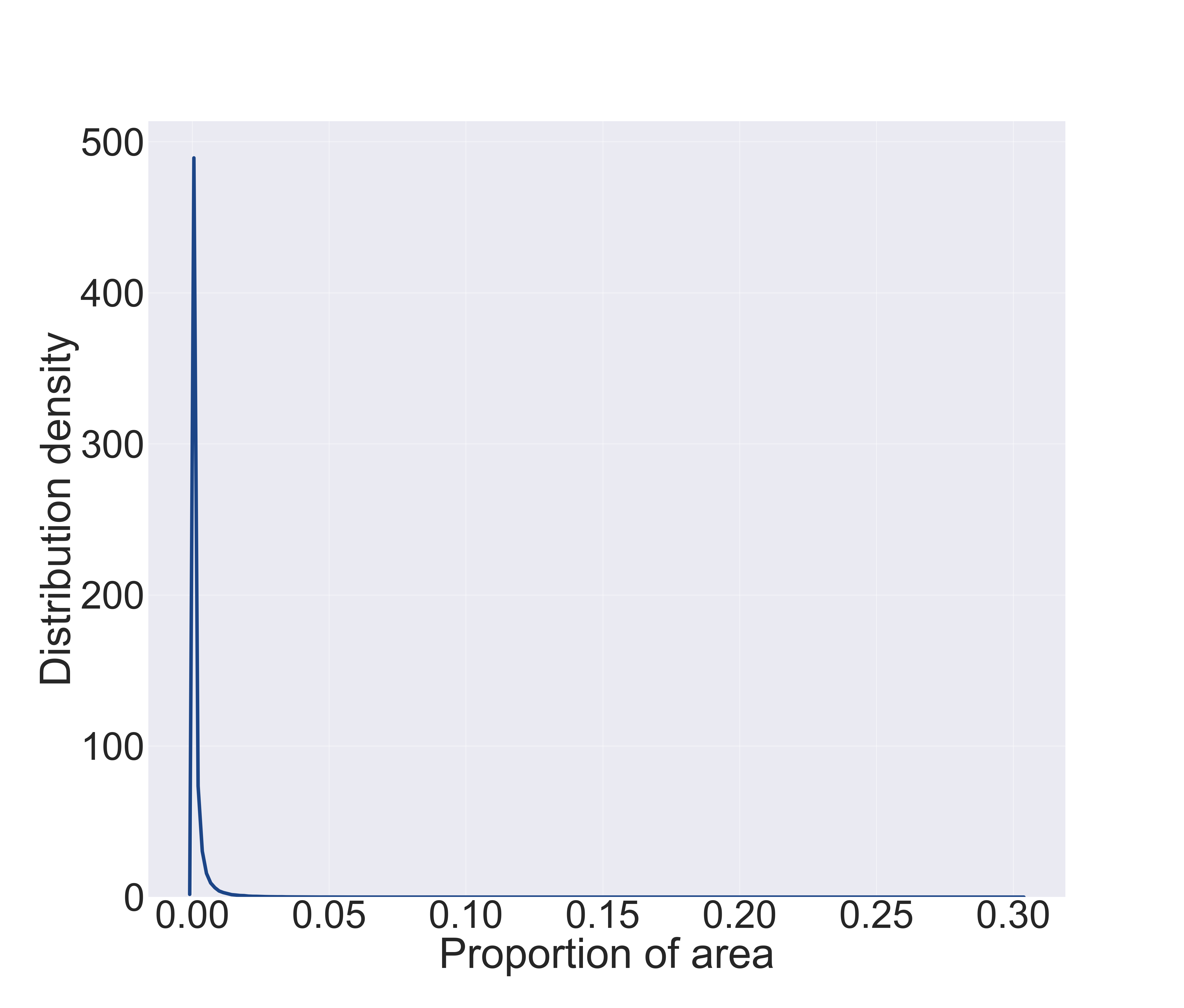}
    }
    \caption{Dataset.}
    \label{info}
\end{figure}

The detailed information of the dataset is visualized in Fig \ref{info}. There are totally 10 classes in this dataset (pedestrian, people, bicycle, car, van, truck, tricycle, awning-tricycle, bus, and motor) as shown in \ref{instances}. Specifically, it can be observed from the \ref{sizes} and \ref{table:size} that 75$\%$ objects are 0.001 times smaller than the image size, indicating the large number of small and tiny objects. Moreover, the label locations \ref{location} indicates that many objects locate in the center of the pictures, which represents the necessity of center-crop data augmentation.

\begin{table}[ht]
\caption{Area of objects}
\centering
\resizebox{0.4\textwidth}{!}{
    \begin{tabular}{|l|l|l|l|l|l|l|}
    \hline
    mean    & std      & min & 25\%    & 50\%     & 75\%     & max      \\
    \hline
    0.001535 & 0.003852 & 0 & 0.00017 & 0.000462 & 0.001342 & 0.302962 \\
    \hline
    \end{tabular}
}
\label{table:size}
\end{table}

\subsection{Hyperparemter Settings}
In order to accelerate the training speed, the input image size is set to be 640 $\times$ 640. We set the batch size to be 128 and training epoch to be 300. We use early stopping strategy to avoid over-fitting, where the patience is set to 15. We use Adam as the optimizer, with an initial learning rate of 0.001. Other detailed parameters are listed in Table \ref{tab:parameters}. The weights of the loss function are set to be 0.5 (object), 0.05 (box) and 0.25 (class) respectively.

There are mainly 5 models in YOLOv5, including YOLOv5n, YOLOv5s, YOLOv5m, YOLOv5l, YOLOv5x. The depth and width of the models increase in sequence while other structures stay the same. The larger the model, the more precise the result. However, in order to accelerate the training speed, we choose to use YOLOv5s during the experiment, with a depth and width of 0.33 and 0.50 respectively. 
 
Data augmentation applied in YOLOv5 contains Mosaic, Copy paste, Random affine, MixUp, HSV augmentation and Cutout as listed in Table \ref{tab:parameters}. We also adopt center crop during this experiment. The height and width of center crop is set to be half of the original image size. It has been observed that center crop is able to improve the overall performance of the model.

\begin{table}[ht]
\caption{Parameter Settings.}
\centering
\resizebox{0.4\textwidth}{!}{
    \begin{tabular}{|l|l|l|l|l|l|l|l|}
    \hline
    hsv\_h & hsv\_s & hsv\_v & degrees & scale & mosaic & mixup & copy\_paste \\
    \hline
    0.4    & 0.3    & 0.5    & 0.2     & 0.4   & 1      & 0.2   & 0.1        \\
    \hline
    \end{tabular}
}
\label{tab:parameters}
\end{table}
        
	Some default anchors are predefined for coco data sets. Before the training starts, annotation details in the dataset will be examined automatically and the most suitable recall rate for the default anchor will be computed. If the optimal recall rate equals or exceeds 0.98, it is not necessary to update the anchor frame. However, YOLOv5 will recalculate the anchors if the optimal recall rate falls below 0.98. During this experiment, 4 groups of anchors of 4 prediction heads are listed in Table\ref{tab:anchors}. Each group is applied for different sizes of feature maps. Specifically, there are 3 pairs of anchors in each group for a single ground truth. Therefore, the overall number of anchors is 4 $\times$ 3=12.

        \begin{table}[ht]
        \caption{Anchor sizes for prediction heads of HIC-YOLOv5.}
        \centering
        \resizebox{0.4\textwidth}{!}{
            \begin{tabular}{|l|l|}
            \hline
            Detection head & Anchor frame size  \\
            \hline
            Tiny           & {[}2.9434,4.0435{]}, {[}3.8626,8.5592{]}, {[}6.8534, 5.9391{]} \\
            \hline
            Small          & {[}10,13{]}, {[}16,30{]}, {[}33,23{]}                          \\
            \hline
            Medium         & {[}30,61{]}, {[}62,45{]}, {[}59,119{]}                         \\
            \hline
            Large          & {[}116,90{]}, {[}156,198{]}, {[}373,326{]}           \\
            \hline
            \end{tabular}
        }
        \label{tab:anchors}
        \end{table}
\subsection{Evaluation criterion}
The common criteria used to evaluate the performance of an object detection algorithm include IoU, Precision, Recall and mAP. The detailed definitions are listed below.

\subsubsection{IoU}
The Intersection over Union (IoU) is calculated by taking the overlap area between the predicted region (A) and the actual ground truth (B) and dividing it by the combined area of the two. The formula can be expressed as     \begin{equation}
    I o U=\frac{A \bigcap B}{A \bigcup B}
    \end{equation}
    
    The value of IoU ranges from 0 to 1. The larger the value, the more precise the model. Particularly, a lower numerator value indicates that the prediction failed to accurately predict the ground truth region. On the other hand, a higher denominator value indicates a larger predicted region, resulting in a lower IoU value.

\subsubsection{Precision}
Precision represents the proportion of samples predicted correctly in the set of samples predicted positively. It can be expressed as \begin{equation}
\text { Precision }=\frac{\text { True positives }}{\text { True positives }+ \text { False positives }}
\end{equation}

\subsubsection{Recall}
Recall represents the proportion of samples that are actually positive and predicted to be correct. It can be expressed as \begin{equation}
\text { Recall }=\frac{\text { True positives }}{\text { True positives }+ \text { False negatives }}
\end{equation}

\subsubsection{mAP}
The Average Precision (AP) is a measure of the Precision scores at different thresholds along the Precision-Recall (PR) curve, and is calculated as a weighted mean. Mean Average Precision (mAP) is the mean values of the AP for all classes. Specifically, mAP@0.5 represents the mAP when IoU is 0.5, mAP@[.5:.95] is the mean mAP when IoU ranges from 0.5 to 0.95.

\subsection{Experimental Results}
    In the conducted experiment, the VisDrone-2019 dataset was utilized to assess the performance of the improved model. After comparing the experimental outcomes between YOLOv5s and the improved algorithm, it can be inferred that our algorithm outperforms YOLOv5s in detecting small targets.
        \begin{table}[ht]
        \caption{Comparison of algorithms on VisDrone2019 dataset.}
        \centering
        \begin{tabular}{|l|l|l|l|}
        \hline
        Method           & Dataset & mAP@.5  & mAP@{[}.5:.95{]} \\
        \hline
        YOLOv5 & Test    & 27.57 & 14.43          \\
        \hline
        Shang et al. \cite{result1}& Test    & 36.4    & 20.1             \\
        \hline
        Liu et al. \cite{result2}  & Test    & 35.3    & 20               \\
        \hline
        Liu et al. \cite{result3}  & Test    & 34.3    & 18.2             \\
        \hline
        Ding et al. \cite{result4} & Val     & 42.9    & 24.6             \\
        \hline
        HIC-YOLOv5       & Test    & 36.95   & 20.85            \\
        \hline
        HIC-YOLOv5       & Val     & 44.31   & 25.95           \\
        \hline
        \end{tabular}
        \label{tab:compare}
        \end{table}

\begin{figure}
            \centering
            \subfloat[YOLOv5]{
                \label{yolov5}\includegraphics[width=0.2\textwidth]{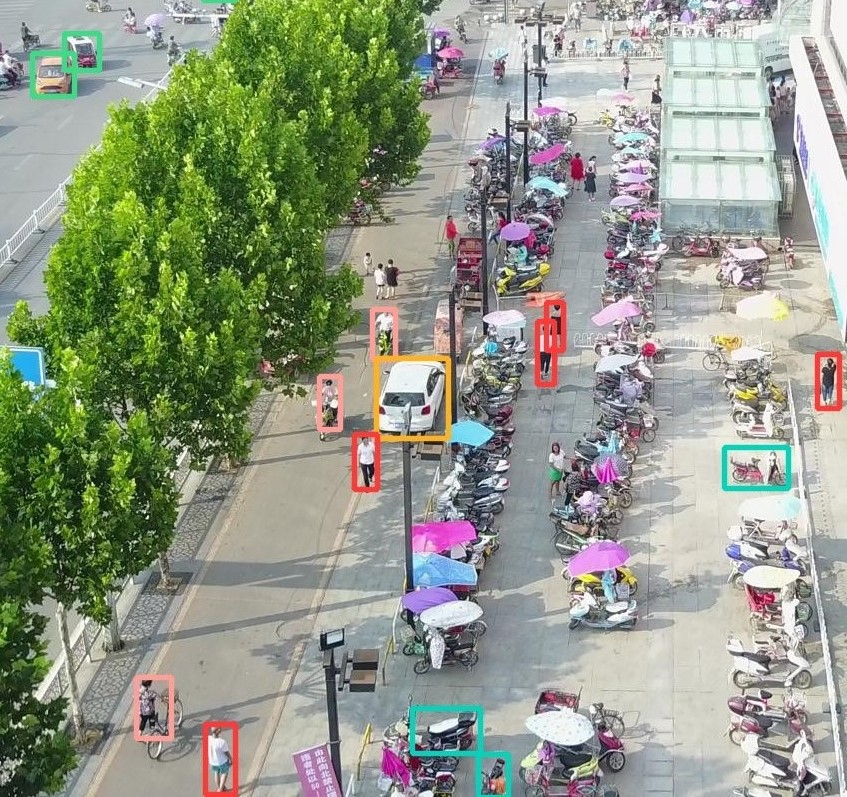}
            }
            \subfloat[HIC-YOLOv5]{
                \label{HIC-YOLOv5}\includegraphics[width=0.2\textwidth]{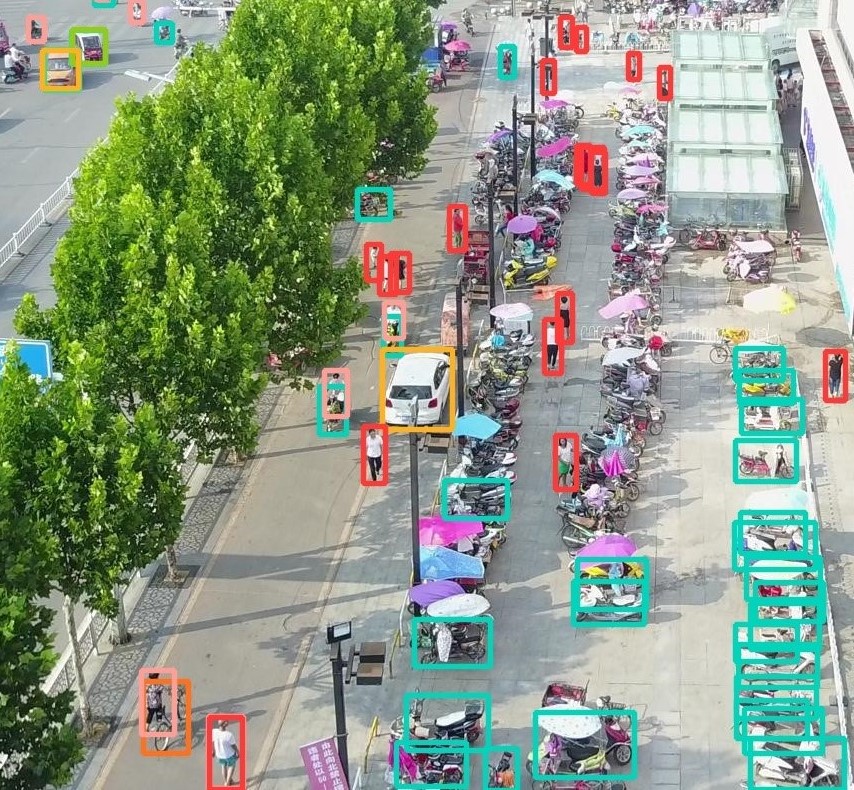}
            }

            \subfloat{
                \label{Legend}\includegraphics[width=0.2\textwidth]{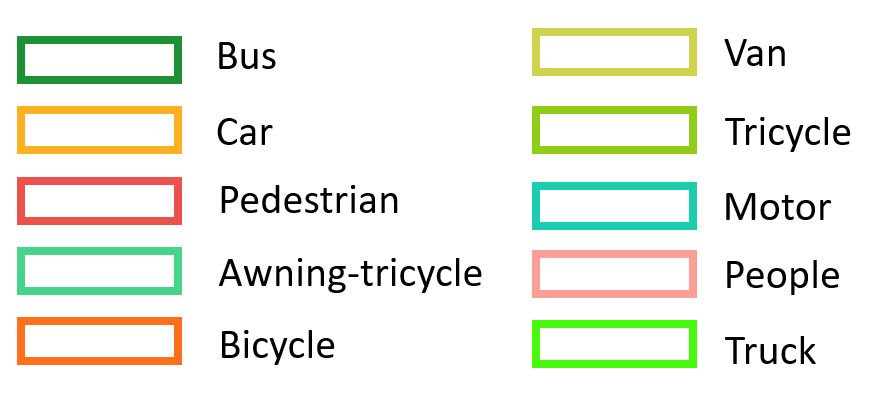}
            }
            \caption{Comparison of detection effect between YOLOv5s and HIC-YOLOv5.}
            \label{performance}
        \end{figure}

	From Table \ref{tab:compare}, we can see that compared with the YOLOv5s model, the mAP@[.5:.95] has been improved by 6.42$\%$ and mAP@0.5 has been improved by 9.38$\%$. Precision and Recall has been improved by 10.29$\%$ and 6.97$\%$ respectively. The small object detection head greatly helps to retain the features of small objects. Additionally, Involution effectively amplifies the channel information, while the CBAM block selectively emphasizes crucial features during their extraction within the backbone. The detection effect between YOLOv5s and HIC-YOLOv5 is shown in Fig \ref{performance}. It visually indicates that more small objects can be detected when using the improved method.

	We also compared our improved model with other algorithms tested on VisDrone2019. The results are shown in Table \ref{tab:compare}. It can be seen that our proposed model has greater performance compared to other detection models. For instance, \cite{result2} applies recursive gated convolution to make greater spatial interaction, but it can lead to the loss of channel information due to 1 $\times$ 1 convolution layers. Compared with \cite{result2}, we adopt involution block to enhance the channel information of PANet, thus improving the performance of small object detection. \cite{10.3389/fpls.2022.1091655} adds a transformer layer at the end of the backbone, which has the shortcoming of large computation cost. Instead, we apply a lightweight CBAM block, which decreases the training time and computational cost. The number of layers, parameters and  gradients of \cite{10.3389/fpls.2022.1091655} and YOLOv5+CBAM are listed in TABLE \ref{tab:computation_compare}. The mAP is incomparable since \cite{10.3389/fpls.2022.1091655} uses another different dataset.
        \begin{table}[ht]
        \caption{Computational comparison of Wang et al.\cite{10.3389/fpls.2022.1091655} and YOLOv5+CBAM.}
        \centering
        \begin{tabular}{|l|l|l|l|}
        \hline
        Model       & Layers & Parameters &  Gradients\\
        \hline
        Wang et al. \cite{10.3389/fpls.2022.1091655} & 297              & 14580167             & 14580167       \\
        \hline
        YOLOv5+cbam & 289              & 8391641              & 8391641   \\
        \hline
        \end{tabular}
        \label{tab:computation_compare}
        \end{table}

\subsection{Ablation Study}
We conducted several experiments to study the effect of three modifications: additional prediction head, involution block and CBAM. The results of ablation study are shown in Table\ref{tab:ablation}. It can be observed that the fourth prediction head makes best contribution to the performance of the model, which improved mAP@.5 and mAP@[.5:.95] by 8.31$\%$ and 5.51$\%$ respectively. Instead, the single block of CBAM and Involution could not improve the model without the help of fourth prediction head. We speculate that it is because there are so many small objects in VisDrone2019 dataset that the single block of CBAM and Involution can not perform well if these small objects can not be detected first. Based on adding the fourth prediction head, the involution block also has great improvement on the model, with 0.66$\%$ and 0.57$\%$ increase of mAP@.5 and mAP@[.5:.95]. Additionally, mAP@.5 and mAP@[.5:.95] are improved by 0.41$\%$ and 0.34$\%$ respectively.

    \begin{table}[ht]
    \caption{Ablation study.}
    \centering
    \resizebox{0.4\textwidth}{!}{
        \begin{tabular}{|l|l|l|l|l|}
        \hline
        Model               & P     & R     & mAP@.5 & mAP@{[}.5:.95{]} \\
        \hline
        Baseline            & 37.59 & 31.52 & 27.57  & 14.43            \\
        \hline
        +SODH                  & 46.18 & 37.89 & 35.88  & 19.94            \\
        \hline
        +cbam                & 36.13 & 28.16 & 24.2   & 11.92            \\
        \hline
        +involution          & 35.49 & 31.13 & 26.76  & 13.8             \\
        \hline
        +SODH+cbam             & 44.5  & 35.35 & 34.53  & 19.03            \\
        \hline
        +SODH+involution      & 46.48 & 37.91 & 36.54  & 20.51            \\
        \hline
        +SODH+involution+cbam & 47.88 & 38.49 & 36.95  & 20.85           \\
        \hline
        \end{tabular}
    }
    \label{tab:ablation}
    \end{table}

\section{CONCLUSIONS}

In this paper, an improved YOLOv5 algorithm HIC-YOLOv5 has been proposed, aiming to improve the performance of small and tiny object detection. There are three main contributions in this paper and the experimental results has proved the effectiveness of our methodology. Firstly, an additional prediction head for small objects is added so that the higher-resolution feature maps could be directly used to detect small targets. Secondly, we adopt an involution block between the backbone and neck, thus increasing channel information of the feature map. Furthermore, we also apply an attention mechanism named CBAM at the end of the backbone to decrease the computation cost and emphasize the important information in both channel and spatial domain. Additionally, data augmentation such as center crop is also applied apart from the original data augmentation methods in YOLOv5. Therefore, the improved YOLOv5 is able to increase the accuracy of detecting small and tiny objects.

\addtolength{\textheight}{-12cm}   








\end{document}